\documentclass[manuscript]{acmart}

\AtBeginDocument{%
  }

\setcopyright{acmlicensed}
\copyrightyear{2026}
\acmYear{2026}
\acmDOI{}
\acmConference[CHI '26 Workshop]{Understanding and Engaging Critical Resistance to AI in Education}{April 13--17, 2026}{Barcelona, Spain}
\acmISBN{}

\acmSubmissionID{1210}

\usepackage{booktabs} 
\usepackage{caption}  
\usepackage{tabularx} 
\usepackage[T1]{fontenc}
\usepackage{enumitem}
\usepackage{graphicx}
\usepackage{epstopdf}
\usepackage{multirow}
\usepackage{subcaption}
\usepackage{xcolor}
\usepackage{siunitx,threeparttable}
\usepackage{tcolorbox}
\tcbuselibrary{skins}
\usepackage{hyperref}
\usepackage{makecell} 
\usepackage{cuted}

\newcommand{\yr}[1]{\textcolor{black}{#1}}
\newcommand{\inj}[1]{\textcolor{black}{#1}}

\begin{document}

\title{When Saying No Makes Better Videos: Designing Dual Gatekeeping for Pedagogically Grounded AI Content Creation}

\author{Yearim Kim}
\authornote{Both authors contributed equally to this research.}
\email{yerim1656@snu.ac.kr}
\affiliation{%
  \institution{Seoul National University}
  \city{Seoul}
  \country{Republic of Korea}
}

\author{Injun Baek}
\authornotemark[1]
\email{jjune1416@snu.ac.kr}
\affiliation{%
  \institution{Seoul National University,}
  \institution{Samsung Electronics}
  \country{Republic of Korea}
}

\author{Nojun Kwak}
\authornote{Corresponding author}
\affiliation{%
  \institution{Seoul National University}
  \city{Seoul}
  \country{Republic of Korea}
}
\email{nojunk@snu.ac.kr}

\renewcommand{\shortauthors}{Kim et al.}

\begin{abstract}
To prevent the adoption of aesthetically polished but pedagogically flawed AI content, we \inj{study} a video authoring pipeline featuring two layers of \textbf{structured refusal}.
The first layer empowers educators to iteratively reshape AI scripts based on multimedia learning theory, while the second employs automated metrics to flag violations in instructional coherence and narrative-visual synchronization.
While neither layer is exhaustive, their synergy ensures that principled resistance--the act of deferring AI output until it meets rigorous standards--becomes a catalyst for higher quality.
\inj{Evaluation combining a study with 23 educators across 3 topics and automated metrics across 7 topics} drawn from established science and philosophy curricula shows that both layers independently improve the same instructional dimensions, suggesting that thoughtful resistance and generative AI are not opposites but partners.
\end{abstract}






\maketitle

\section{Introduction}
While modern AI models~\cite{google2024veo, openai2024sora} synthesizes professional-looking educational videos in a minute, surface-level polish does not guarantee pedagogical rigor. 
Current video generation pipelines often prioritize visual appeal over instructional essentials, such as precise temporal alignment of narration or strategic sequencing of prerequisite concepts.
Consequently, outputs are frequently optimized for \textit{looking good}, rather than \textit{teaching well}.
This gap matters as educators are increasingly expected to adopt AI-generated materials with minimal intervention. The pressure toward seamless, friction-free adoption treats any slowdown as inefficiency---a stance that risks reducing the educator's role from professional decision-maker to passive consumer. 
We argue, however, that pedagogical friction is not a hurdle to be eliminated but a site of professional accountability.
Moments of deliberate hesitation--whether an educator questioning a script's logical flow or an algorithm flagging a narrative truncation--are precisely where instructional quality is forged.

\begin{table}[h]
\vspace{-10pt}
\caption{Mayer's 12 CTML principles~\cite{mayer2009} for reducing extraneous, managing essential, and fostering generative processing.}
\vspace{-10pt}
\label{tab:ctml}
\centering
\small
\begin{tabular}{llllll}
\toprule
Coherence & Signaling & Redundancy & Spatial Contiguity & Temporal Contiguity & Segmenting \\
Pre-training & Modality & Multimedia & Personalization & Voice & Image \\
\bottomrule
\end{tabular}
\vspace{-12pt}
\end{table}

We conceptualize this approach as \textit{principled resistance}: deliberate, theory-grounded pushback against AI outputs that fail pedagogical standards.
Rooted in Mayer's Cognitive Theory of Multimedia Learning~(CTML)~\cite{mayer2009}--a framework of 12 empirically validated principles for effective multimedia instruction (Table~\ref{tab:ctml})--we developed the PedaCo (Pedagogical Co-creation), a human-AI collaborative system that operationalizes this resistance.
PedaCo integrates two complementary gatekeeping mechanisms: reviewing scripts through CTML-informed criteria, and an automated metric evaluating finished videos against computationally measurable CTML dimensions.
This paper presents the design rationale behind this dual approach, summarizes converging evidence from both human and computational evaluations, and poses open questions about how educational AI systems should balance human agency with automated safeguards.

\section{Two Layers of Resistance}
In our framework, principled resistance takes three concrete forms: \textbf{rejecting} (requesting regeneration), \textbf{revising} (manual editing), and \textbf{overriding} (vetoing automated flags).
These are not ad hoc reactions but norm-driven decisions grounded in CTML principles.
The framework rests on a simple observation: educators and algorithms are good at catching different kinds of problems:
\yr{while human educators excel at identifying nuanced pedagogical mismatches--such as content being too advanced for a target audience--algorithms provide precise, high-resolution verification of structural integrity, such as identifying temporal misalignments between narration and visuals.}
Building both checkpoints into the same pipeline creates overlapping coverage that neither could achieve alone.

\subsection{Layer 1: Human Educator Review at the Script Stage}
The first layer intervenes before any video is rendered (Figure~\ref{fig:system} left). \inj{The educator begins by inputting learning content and configuring which CTML principles the system should enforce.} An LLM~\cite{team2023gemini} then generates an initial script, which passes through a structured review cycle. An AI reviewer---itself prompted with CTML principles---produces feedback organized by principle, identifying potential violations rather than definitive judgments (e.g., ``Scene 3 introduces technical terms without prior explanation, which may conflict with the Pre-training principle''). The human educator then decides what to accept, what to revise manually, and what to regenerate. \inj{This review loop can be repeated until the educator is satisfied with the script.}

\yr{This design choice to review at the script level is deliberate.
Textual revisions are computationally and laboriously efficient, whereas pedagogical errors baked into a rendered visual narrative are nearly impossible to correct post-synthesis.}
By placing the human checkpoint at this intermediate stage, we make pedagogical critique economically viable.
\yr{This ensures the system remains advisory, preserving the educator’s professional authority to say `no' to AI suggestions based on their specific curricular context and pedagogical style.}


\subsection{Layer 2: Automated Metric After Video Synthesis}
\begin{figure}[t]
\vspace{-10pt}
  \centering
    \includegraphics[width=.95\linewidth]{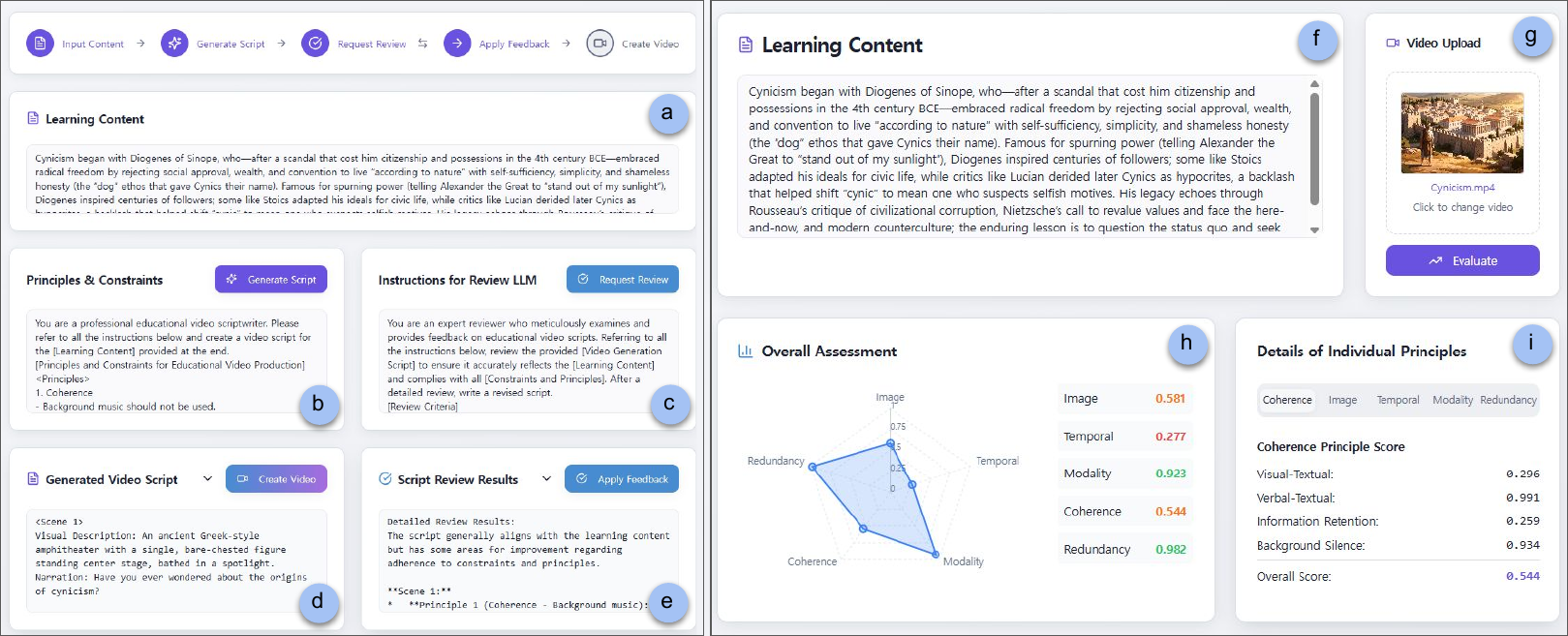}
\vspace{-5pt}
\caption{\textbf{The Dual Gatekeeping Interface.} The left panel \textbf{(Layer 1: Script Level)} generates an initial script ($d$) from learning content ($a$) and generation principles ($b$), then scaffolds the educator's revision by providing AI critiques and revised drafts ($e$) based on review constraints ($c$). The right panel \textbf{(Layer 2: Video Level)} visualizes invisible pedagogical quality via automated metrics ($h, i$), allowing users to assess the alignment between the final video ($g$) and the learning content ($f$).}
\label{fig:system}
\vspace{-20pt}
\end{figure}

\yr{The second layer performs a post-synthesis evaluation of the finalized video through a composite metric (Figure~\ref{fig:system} right).
We automate the assessment of five dimensions: \textbf{coherence}, \textbf{redundancy}, \textbf{temporal contiguity}, \textbf{modality}, and \textbf{image quality}.} \inj{The educator reviews the principle-level scores and decides whether to accept the video or return to the script stage for targeted revision.}
\yr{This boundary between human and automated resistance is a strategic design decision.
While dimensions like temporal synchronization are amenable to reliable computational measurement, others—such as Personalization (tone)—require a deep understanding of curricular structures and learner psychology that currently remains the sole domain of the human expert.
By automating only where algorithmic feasibility aligns with pedagogical necessity, we create a robust safety net that prevents technical regressions without marginalizing human judgment.}

\section{Evidence from Two Evaluations}
\yr{We conducted a multi-method evaluation to assess the efficacy of the PedaCo framework, combining a human-centric study with educators~(3.1) and an algorithmic assessment via automated metrics~(3.2). Together, these evaluations provide converging evidence for the value of dual-layered resistance.}

\subsection{What Educators Found}
We conducted a within-subject study with 23 educators—who were priorly briefed on the CTML principles—directly used our system to generate and evaluate videos. 
Guided by the system, participants simulated  with the pre-generated videos by inputting raw learning content and iteratively refining the AI-generated scripts with the system's CTML feedback.
Then, they compared these videos outcomes with baseline videos generated without CTML guidelines.
The topics include three different cognitive demands: causal reasoning, abstract concepts, and procedural knowledge.
Participants rated each condition on 13 items covering all 12 CTML principles and overall instructional validity.

\yr{The review-based approach yielded statistically significant improvements across every principle ($p < .05$, Wilcoxon signed-rank).}
The mean rating rose from 3.07 to 3.86 on a 5-point scale (+0.79, $p < .01$), \yr{with the most pronounced gains observed in content organization: prerequisite sequencing ($+0.86$), irrelevant material removal ($+0.84$), and overall instructional validity ($+0.96$)ove.}
\inj{Nearly} all effects were large ($r \geq .64$, computed as $Z/\sqrt{N}$)\inj{, with one exception (redundancy, $r = .42$, medium),} and consistent across gender and experience level.

Notably, educators did not perceive the review process as slowing them down.
They rated production efficiency at 4.26/5 and the validity of the CTML-based guidance at 4.04/5 (with remarkably low variance, $SD = 0.62$).
One participant captured the tension well: the iterative process was \textit{``quite challenging''} but \inj{ultimately} for producing \textit{``a robust and effective learning tool''} (P23).
Another explicitly requested \textit{``separate functions where teachers can additionally review, edit, and modify''} the AI output (P05)---in other words, more resistance, not less.

\subsection{What the Metrics Found}
\yr{Independently, we applied our automated metrics to a corpus of 14 videos (7 topics $\times$ 2 conditions) drawn from established science and philosophy curricula.}
\yr{The videos were generated with identical structure, isolating the CTML-informed \inj{generation and} review as the sole variable.}
Two of the five metrics showed significant improvement: temporal contiguity (0.294 vs. 0.273, $p = .021$) and coherence (0.729 vs. 0.646, $p = .011$). 
The remaining three showed no significant difference---modality and redundancy scored high in both conditions (near ceiling), and image quality did not differ as both conditions used the same video synthesis model.
We retain these metrics as safety nets: current models perform well on these dimensions, but they serve as guardrails against future model regressions or hallucination-induced failures.

\subsection{Where the Two Evaluations Agree}
\yr{The most compelling evidence for our dual-layered approach is the high degree of convergence between subjective ratings and objective metrics.
Despite being conducted independently with different samples and instruments, both evaluations identified \textbf{coherence} and \textbf{temporal alignment} as the dimensions most significantly enhanced by the PedaCo pipeline.
Educators ranked coherence among the top three improvements, matching the statistically significant gains identified by the automated system.
This triangulation suggests that the two gatekeeping layers are not merely redundant but provide complementary verification of the same underlying instructional quality.
}

\section{Discussion: Reframing Resistance in  ducational AI}
\yr{The PedaCo framework offers a theory-grounded perspective on where the boundaries of Generative AI should be drawn in educational settings.
By operationalizing "principled resistance" through CTML, we illustrate how intentional friction can sustain \textbf{pedagogical integrity} without marginalizing the professional authority of educators.
Our findings suggest that "productive friction" is most effective when it is: (a) theoretically grounded rather than intuitive; (b) strategically embedded upstream at the script stage; and (c) hybridized across human and computational agents.}

\yr{However, this dual-gatekeeping approach surfaces three emergent tensions for the workshop to consider:
\begin{itemize}
    \item \textbf{Negotiating Agency}: When automated flags and educator judgments diverge, how should the interface balance algorithmic safeguards with human autonomy?
    \item \textbf{Sustainability of Friction}: While valued in our short-term study, the long-term viability of iterative review in daily classroom preparation remains unknown. We must identify the threshold where productive friction transitions into "friction fatigue."
    \item \textbf{Beyond Proxy Metrics}: Future research must move beyond theoretical proxies to evaluate the direct causal impact of structured resistance on student learning outcomes.
\end{itemize}}
\section{Conclusion}

In this paper, we have argued that educational AI needs structured ways to say ``not yet''---\yr{bridging human pedagogical expertise with computational precision.}
Our PedaCo framework builds principled resistance into the video generation pipeline through educator review at the script stage and automated pedagogical metrics at the video stage.
Early evidence suggests both layers improve the same instructional dimensions, and that educators experience this friction as productive rather than burdensome.
We offer this as one concrete answer to the workshop's animating question: resistance to AI in education \yr{should not be equated with rejection.} 
\yr{Instead,} it can mean building systems that are designed to push back, on principled grounds, until the output is genuinely ready to teach.

\begin{acks}
This work was funded by the Korean Government through the grants from IITP (RS-2021-II211343) and KOCCA (RS-2024-00398320).
\end{acks}

\bibliographystyle{ACM-Reference-Format}
\bibliography{main}

@misc{google2024veo,
  title   = {Veo: Google’s most capable generative video model},
  author  = {Google DeepMind},
  year    = {2024},
  month   = {May},
  howpublished = {\url{https://deepmind.google/models/veo/}},
  note    = {Accessed: 2025-09-29}
}

@inbook{mayer2009, place={Cambridge}, title={Multimedia Principle}, booktitle={Multimedia Learning}, publisher={Cambridge University Press}, author={Mayer, Richard E.}, year={2009}, pages={223–241}}

@article{openai2024sora,
  title={Video generation models as world simulators},
  author={Tim Brooks and Bill Peebles and Connor Holmes and Will DePue and Yufei Guo and Li Jing and David Schnurr and Joe Taylor and Troy Luhman and Eric Luhman and Clarence Ng and Ricky Wang and Aditya Ramesh},
  year={2024},
  url={https://openai.com/research/video-generation-models-as-world-simulators},
}

@article{team2023gemini,
  title={Gemini: a family of highly capable multimodal models},
  author={Team, Gemini and Anil, Rohan and Borgeaud, Sebastian and Alayrac, Jean-Baptiste and Yu, Jiahui and Soricut, Radu and Schalkwyk, Johan and Dai, Andrew M and Hauth, Anja and Millican, Katie and others},
  journal={arXiv preprint arXiv:2312.11805},
  year={2023}
}

\end{document}